\documentclass[10pt,twocolumn,letterpaper]{article}

\usepackage[pagenumbers]{iccv} 

\newcommand{\algabbr}{POD\xspace}

\definecolor{iccvblue}{rgb}{0.21,0.49,0.74}
\usepackage[pagebackref,breaklinks,colorlinks,allcolors=iccvblue]{hyperref}

\def\paperID{15455} 
\def\confName{ICCV}
\def\confYear{2025}

\newcommand{\algname}{POD}

\title{Predict-Optimize-Distill: A Self-Improving Cycle for 4D Object Understanding}

\author{
Mingxuan Wu* \quad
Huang Huang* \quad
Justin Kerr \quad
Chung Min Kim \quad 
\\
Anthony Zhang \quad
Brent Yi \quad
Angjoo Kanazawa \\
\vspace{-0.5em}
\\  
University of California, Berkeley
}
\begin{document}

\twocolumn[{%
\renewcommand\twocolumn[1][]{#1}
\maketitle
\vspace{-2.5em}
\begin{center}
    \begin{minipage}{\linewidth}
        \centering
        \includegraphics[width=\textwidth,trim=3 4 4 4,clip]{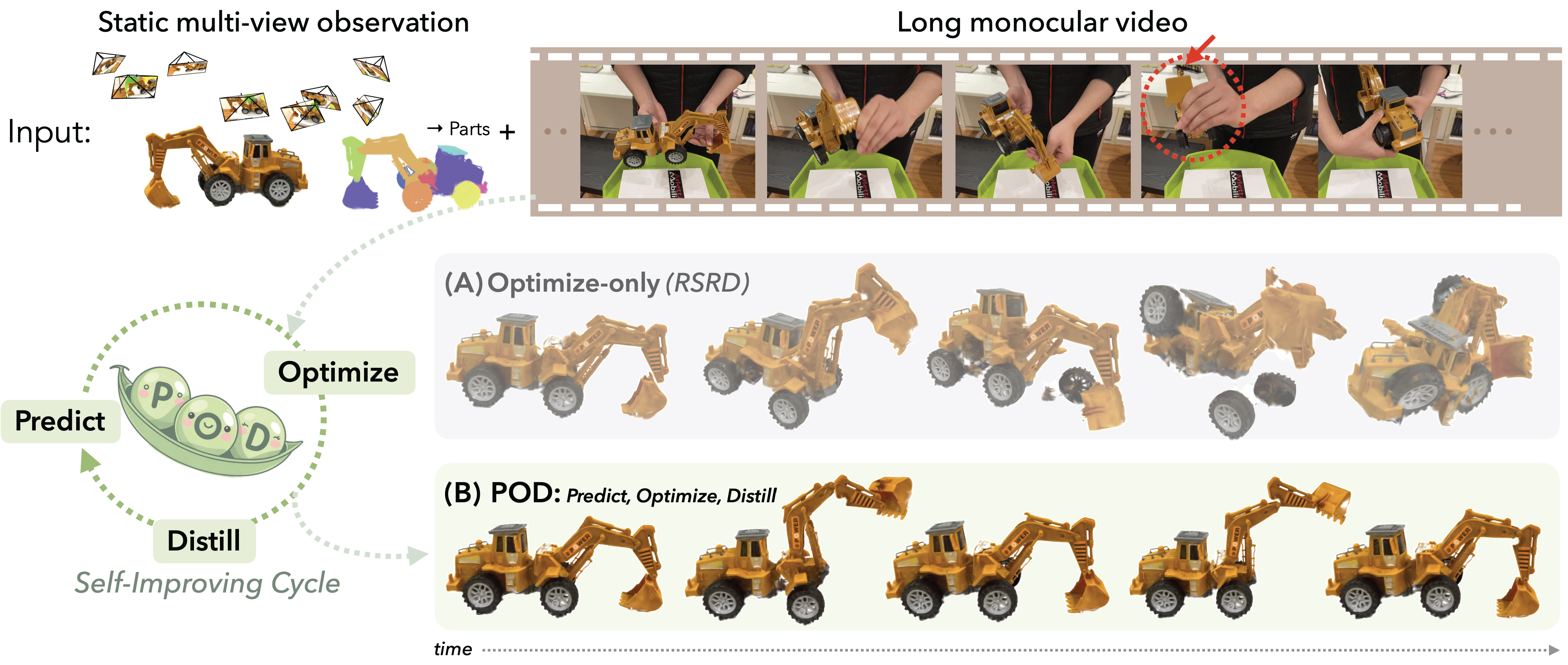}
        \captionof{figure}{\textbf{Predict-Optimize-Distill (\algname{})} takes in a multi-view scan of an object and casually captured long-form human interaction video, and estimates 3D part poses over time.
        (A) Existing optimization-based methods experience failures under heavy occlusion or when incremental frame optimization drifts. (B) In contrast, \algname{} utilizes a cycle consisting of a predictive feed-forward model, an optimization stage, and a self-distillation phase to iteratively improve the object part pose predictions. By training an object part pose prediction model, \algname{} can predict correct part poses even under heavy occlusions in the observations (red circle).
        }
        \label{fig:splash}
    \end{minipage}
\end{center}
}]

\begin{abstract}

Humans can resort to long-form inspection to build intuition on predicting the 3D configurations of unseen objects. The more we observe the object motion, the better we get at predicting its 3D state immediately. 
Existing systems either optimize underlying representations from multi-view observations \emph{or} train a feed-forward predictor from supervised datasets. We introduce Predict-Optimize-Distill (\algabbr{}), a self-improving framework that interleaves prediction and optimization in a mutually reinforcing cycle to achieve better 4D object understanding with increasing observation time. Given a multi-view object scan and a long-form monocular video of human-object interaction, POD iteratively trains a neural network to \textbf{predict} local part poses from RGB frames, uses this predictor to initialize a global \textbf{optimization} which refines output poses through inverse rendering, then finally \textbf{distills} the results of optimization back into the model by generating synthetic self-labeled training data from novel viewpoints. Each iteration improves both the predictive model and the optimized motion trajectory, creating a virtuous cycle that bootstraps its own training data to learn about the pose configurations of an object. We also introduce a quasi-multiview mining strategy for reducing depth ambiguity by leveraging long video. We evaluate POD on 14 real-world and 5 synthetic objects with various joint types, including revolute and prismatic joints as well as multi-body configurations where parts detach or reattach independently. 
\algabbr{} demonstrates significant improvement over a pure optimization baseline which gets stuck in local minima, particularly for longer videos. We also find that \algabbr{}'s performance improves with both video length and successive iterations of the self-improving cycle, highlighting its ability to scale performance with additional observations and looped refinement. See our website at: \url{https://predict-optimize-distill.github.io/pod.github.io}
\end{abstract}    
\vspace{-1em}
\section{Introduction}


Given a familiar object like a pair of scissors, humans can effortlessly recognize the 3D configuration of its parts from a single glance---a hallmark of fast, automatic System 1 thinking. We build this intuition through slower System 2 processes: interacting with unfamiliar objects, manipulating them to uncover their structure, and gradually optimizing our internal models. Over time, this interplay becomes self-improving: fast predictions provide better starting points for optimization, which in turn leads to better solutions and more accurate future predictions. In contrast, most 4D computer vision systems are either optimization-based---refining structure from multiple observations---or feed-forward---predicting structure directly from static input. Few systems interleave the two in a way that improves over time.




In this work, we introduce Predict-Optimize-Distill (\algabbr{}), a framework that interleaves prediction and optimization in a mutually reinforcing cycle to obtain better 4D object understanding with longer observation of the object. 
Specifically, the scenario mimics an embodied setting where an agent can first observe the object at rest from multiple views, then is presented with a long form ($\sim$15-30 seconds) monocular video where a human demonstrates repeated motions of the object by reorienting it in their hands. The goal is then to reconstruct the 3D pose of parts over time as well as the camera to object transform.
Reconstructing the 3D configuration of objects with moveable parts from casually captured monocular videos is a challenge due to depth ambiguity, object self-occlusion, and hand-object occlusion. These factors cause pure optimization approaches like RSRD~\cite{rsrd} to easily fall into irrecoverable local minima. Further, longer videos make it \textit{harder} for such systems due to drifting issues (Fig.~\ref{fig:splash}), 
while ideally a data driven approach should scale performance with observation length. On the other hand, purely predictive approaches also fall short due to the lack of large-scale 4D datasets covering the full diversity of man-made objects and their articulations.

\algabbr{}'s key insight is to bootstrap both training data and a predictive model in a self-supervised way; we iteratively train an image-conditioned feed-forward model to predict the local poses of the object's parts from a single frame, then initialize a global optimization with these predictions and refine them with inverse rendering to explain the input video. 
We start with a over-segmented 3D part aware reconstruction~\cite{garfield}.  The refined object configurations after optimization are then used to generate a synthetic training dataset by self-labeling rendered configurations from novel views. 
Critically, by training the pose predictor on rendered synthetic views of the recovered trajectory, we expose it to a diverse set of viewpoints while supervising it with a clean, ambiguity-free ground truth pose signal. 
Each iteration of the cycle initializes the optimization with the latest pose predictor model, making it easier to converge to the correct solution, which in turn improves the quality of the pose predictor serving as the training data. This creates a cycle where prediction and optimization mutually reinforce each other.

During optimization, we minimize a pixel-based projection loss similar to RSRD~\cite{rsrd}, comparing rendered object embeddings to extracted frame embeddings.
We also leverage the fact that longer video sequences often capture different views of the same part configuration at different times, allowing us to mine quasi-multiview supervision. Specifically, \algabbr{} uses its feed-forward model to identify frame pairs with similar local part configurations and jointly optimizes through both views into the same pose parameters. This enforces consistency across views, helping the model resolve depth ambiguity. This process is part of the self-improving cycle---as the predictive model improves, it identifies better quasi-multiview frame pairs, further enhancing supervision and refinement.
In addition we imposes temporal smoothness and soft rigidity priors, which all help improve upon the coherence of the predicted object configurations. 

We evaluate \algabbr{} on 14 real-world objects with a variety of rotary, ball, and prismatic joints, and on 5 synthetic objects for ground-truth pose error evaluation. Our approach significantly outperforms pure optimization methods while showing robust performance even under heavy occlusions and depth ambiguity, accurately reconstructing coherent 3D part configurations from monocular input. 
We also observe that POD’s performance improves with both longer video input and continued self-improvement iterations, highlighting its scalability with increased data and compute.
This improvement is quantitatively measured as a substantial increase in the percentage of correct points (PCP) metric.
\begin{figure*}[t!]
    \centering
    \includegraphics[width=\linewidth,trim=4 4 4 4,clip]{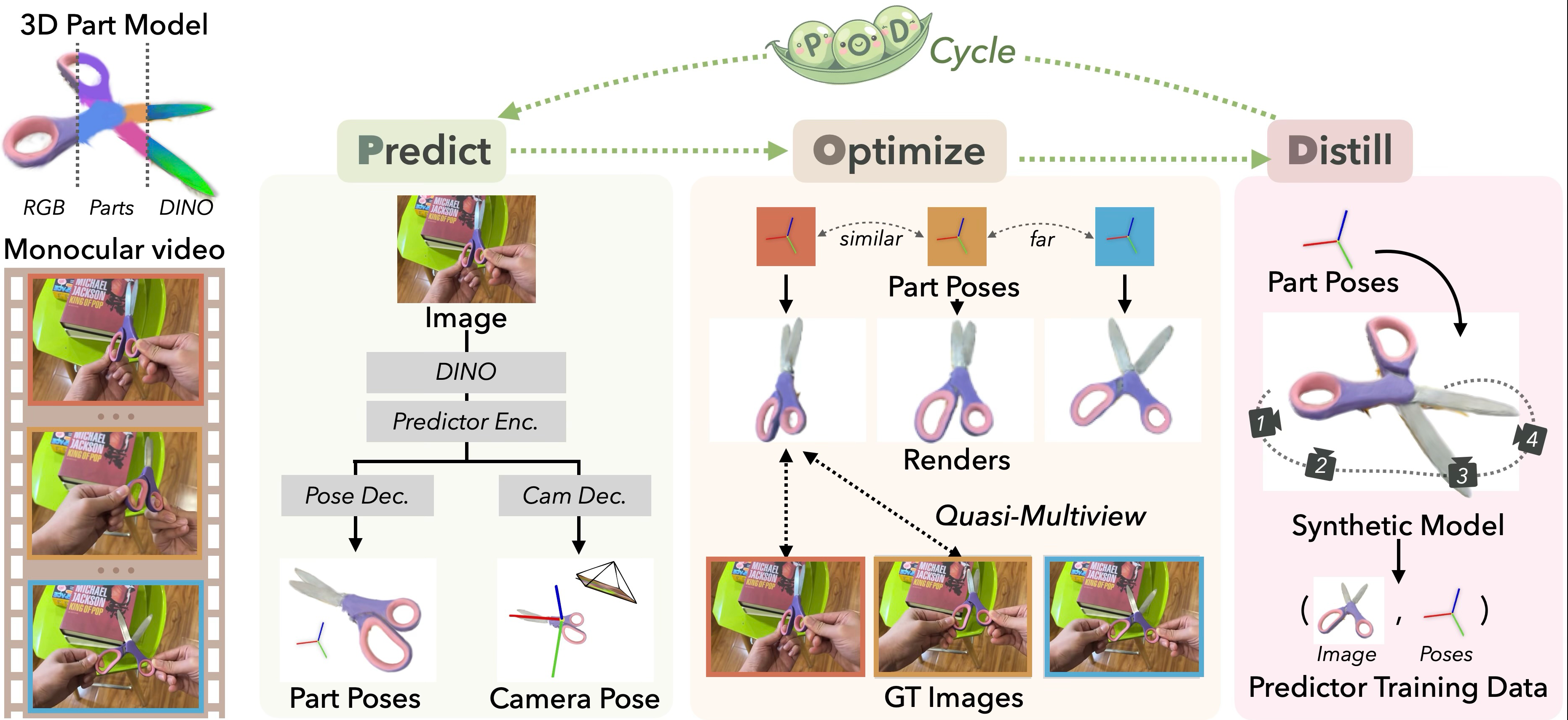}
    \caption{
    \textbf{\algname{} Pipeline:}  \algname{} builds a 3D object model from a multi-view scan of the object using 3D Gaussian Splatting~\cite{3dgs} and GARField~\cite{garfield}. In the \textbf{predict} stage, \algname{} estimates the object's part poses from a monocular video with a feed-forward model conditioned on RGB images. Using these predictions it \textbf{optimizes} poses against the monocular video observations, utilizing quasi-multiview supervision by finding matching frames with similar predicted poses and jointly optimizing them. 
    \algabbr{} then \textbf{distills} the optimized poses back into the predictive model by generating synthetic data from novel views by applying part poses to the object 3D model and rendering RGB observations from different camera poses. 
    }
    \label{fig:pipeline}
\end{figure*}
\section{Related Work}

There is a large body of work on reconstructing arbitrary deforming scenes from monocular videos~\cite{park2021nerfies,tretschk2021non,pumarola2020d,gao2022dynamic,som2024,lei2024mosca}. Here, we focus on methods that recover the 3D state of articulated, interactable objects. 


\vspace{0.5em}\noindent\textbf{Predictive methods}
\looseness=-1 Most existing predictive models leverage predefined 3D template models such as SMPL~\cite{smpl}, FLAME~\cite{flame}, MANO~\cite{mano}, and SMAL~\cite{zuffi} for human bodies, faces, hands, and quadrupeds, respectively. Approaches like human mesh recovery~\cite{kanazawa2018end,goel2023humans,pavlakos2024reconstructing,bite2023rueegg} exploit these templates to directly predict pose parameters from images, benefiting from large-scale motion capture and annotated datasets. Methods focusing on animals without 3D templates~\cite{kanazawa2018learning,kulkarni2020articulation,li2024fauna,kaye2025dualpm,yang2022banmo,yao2022lassie} jointly infer shape parameters---including shape and articulation---and predictive models directly through analysis-by-synthesis from image or video data. However, these techniques primarily rely on inductive biases such as representing shapes and articulations as deformations of a single sphere, combined with kinematic structures and linear blend skinning. In contrast, we introduce a method that bootstraps an instance-specific predictive model directly from long-form monocular video interactions. Unlike existing methods, our approach generalizes beyond purely revolute joints to accommodate more general articulations, including prismatic joints and multi-body configurations where parts detach or reattach independently.

For man-made objects, CAPTRA~\cite{weng2021captra} performs 9DoF tracking of category-level poses from pointcloud streams, requiring known initial per-part poses and object part counts. 
CARTO~\cite{heppert2023carto} learns a predictive model that directly outputs geometry and articulation parameters from stereo images trained on PartNet-Mobility~\cite{xiang2020sapien}. Given RGB-D input, robotics inspired methods~\cite{heppert2022category,jiang2022ditto} reconstruct part-level geometry and estimate articulation models using neural networks trained on 3D articulated object datasets~\cite{wang2019shape2motion,abbatematteo2019learning}. These learning-based approaches struggle with objects outside their training categories. Unlike these methods, \algabbr{} doesn't require curated synthetic datasets with 4D supervision, instead generating its own synthetic 4D data from ongoing observations, creating a Real-to-Sim-to-Real cycle without any a priori part configuration training data. We also operate directly on RGB images without the pointcloud assumption, which requires resolving ambiguities.

\paragraph{Optimization based approaches}
Prior work~\cite{liu2023reart,weng2024neural} has learned articulated object models from RGBD video or pointclouds. In this work we take as input only a monocular video, presenting significant challenges in depth ambiguity. PARIS~\cite{jiayi2023paris}, Weng~\etal~\cite{weng2024neural}, and Art-GS~\cite{liu2025building} also reconstruct articulated objects with NeRF or 3DGS while learning the part assignment and articulation model from multiview images. Critically, these approaches all rely on multi-view observation of the object after each interaction or require simultaneous multi-view video observations~\cite{noguchi2021watch}, which severely limits the usability of such methods on complex object behaviors. In contrast, \algabbr{} operates on monocular long-form videos of object interaction which enables more casual capture of object dynamics and introduces significant challenges of occlusion. Furthermore, we model object motion as set of relative SE(3) transformations, which enable reconstructing prismatic articulation or object parts that move apart. Other approaches learn dynamic motions with highly unconstrained motion fields or neural representations~\cite{gao2022dynamic, park2021hypernerf,wang2023omnimotion}, however these works do not produce transferable predictive models as \algabbr{} does, and also do not scale performance with video length. The work closest to the setting in this paper RSRD~\cite{rsrd}, but as we demonstrate in Figure~\ref{fig:splash}, it is a pure optimization based approach which struggles with long videos, and in addition cannot improve with successive observations as \algabbr{} does.



\paragraph{Self-Improving Cycle}
There is growing interest in methods that continually self-improve by bootstrapping predictive models based on intermediate results~\cite{zelikman2022star}. For instance, SPIN~\cite{kolotouros2019spin} demonstrates such a cycle by supervising human pose predictors  with intermediate optimization results using 2D keypoints. Agent-to-Sim~\cite{yang2024ats} similarly uses synthesized global poses derived from template reconstructions, though it does not iteratively refine beyond this initial step.
Inspired by these approaches, \algabbr{} extends the iterative self-improvement paradigm to general articulated, man-made objects, continually bootstrapping instance-level predictive models through repeated cycles of prediction, optimization, and distillation. To further resolve the ambiguities inherent in monocular video, we leverage current predictions to mine quasi-multiview supervision from long-form video observations. As the predictive model improves, this additional supervision becomes increasingly effective, creating a virtuous cycle of continual improvement.

\begin{figure}[t!]
    \centering
    \includegraphics[width=\linewidth]{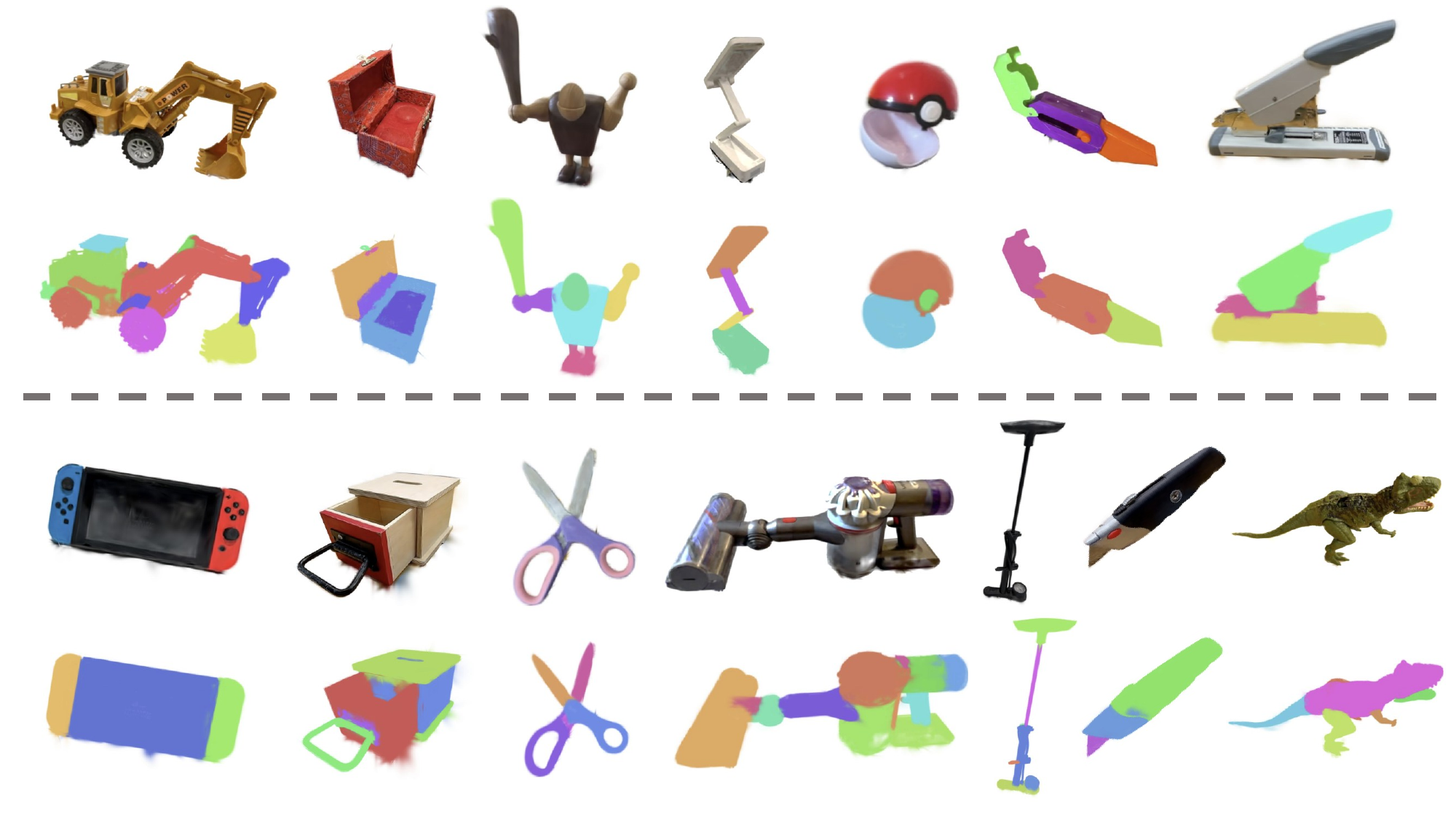}
    \caption{
    \textbf{Example real objects} used for qualitative evaluation. (Top) RGB renderings, (Bottom) part segmentations. Please see supplemental for more. 
    } \label{fig:real-data-groups}
    \vspace{-1em}
\end{figure}
\section{The Predict-Optimize-Distill Cycle}

\algabbr{} takes as input a multi-view object scan with part segmentation~\cite{garfield} and a video of a person manipulating the object's parts. The goal is to obtain a 4D reconstruction of the object, specifically per-frame 3D local and global pose parameters defined below.
\algabbr{}'s cycle contains a \textit{predictive model}, an end-to-end image conditioned neural network that predicts the poses of parts and camera-to-object transformation from a RGB image, an \textit{optimization stage} which improves upon the predicted poses 
by minimizing pixel losses compared to the input video through inverse rendering
and a \textit{distillation phase} where optimized poses are mined for synthetic ground truth to continue fine-tuning the pose prediction model. Since the optimization stage receives increasingly better initializations, it produces more accurate refinements, which in turn improve the predictive model. As a result, each stage enhances the next, leading to robust 4D reconstruction even under heavy hand-object interaction or object self-occlusion—scenarios where optimization alone fails. 

\noindent\textbf{3D Template Model}
Similar to how a person might inspect a new object by turning it around in their hands to understand its shape and structure, \algabbr{} first builds a 3D Gaussian Splat (3DGS)~\cite{3dgs} from the object scan and decomposes it into parts via GARField~\cite{garfield} by selecting valid scale for grouping. It then embeds DINOv2~\cite{oquab2023dinov2} features into the 3DGS following~\cite{rsrd, qin2023langsplat,featuresplatting,kobayashi2022distilledfeaturefields}. Similar to RSRD~\cite{rsrd}, we parameterize part configurations by defining an object  frame at the object's centroid and representing part poses relative to it. Each part has a local frame at its centroid. The object's full configuration is specified by individual part-to-object transforms along with an object-to-camera transform. Specifically, if an object has $P$ parts, each part $p_i$ is transformed relative to the object frame as $T^{obj}_{p_i} \in SE(3)$, and the object's global pose is estimated relative to the input camera as $T_{obj}^{cam} \in SE(3)$.  We define the set of part transformations as $T^{obj}_{parts} = \{T^{obj}_{p_1}, T^{obj}_{p_2}, \dots, T^{obj}_{p_P} \}$, which serve as the local pose parameters. This forms a one-level kinematic hierarchy, enabling both rotary and prismatic joints as well as arbitrary part motions. This template is used throughout the method for both prediction and optimization.
\begin{figure}[t!]
    \centering
    \includegraphics[width=\linewidth]{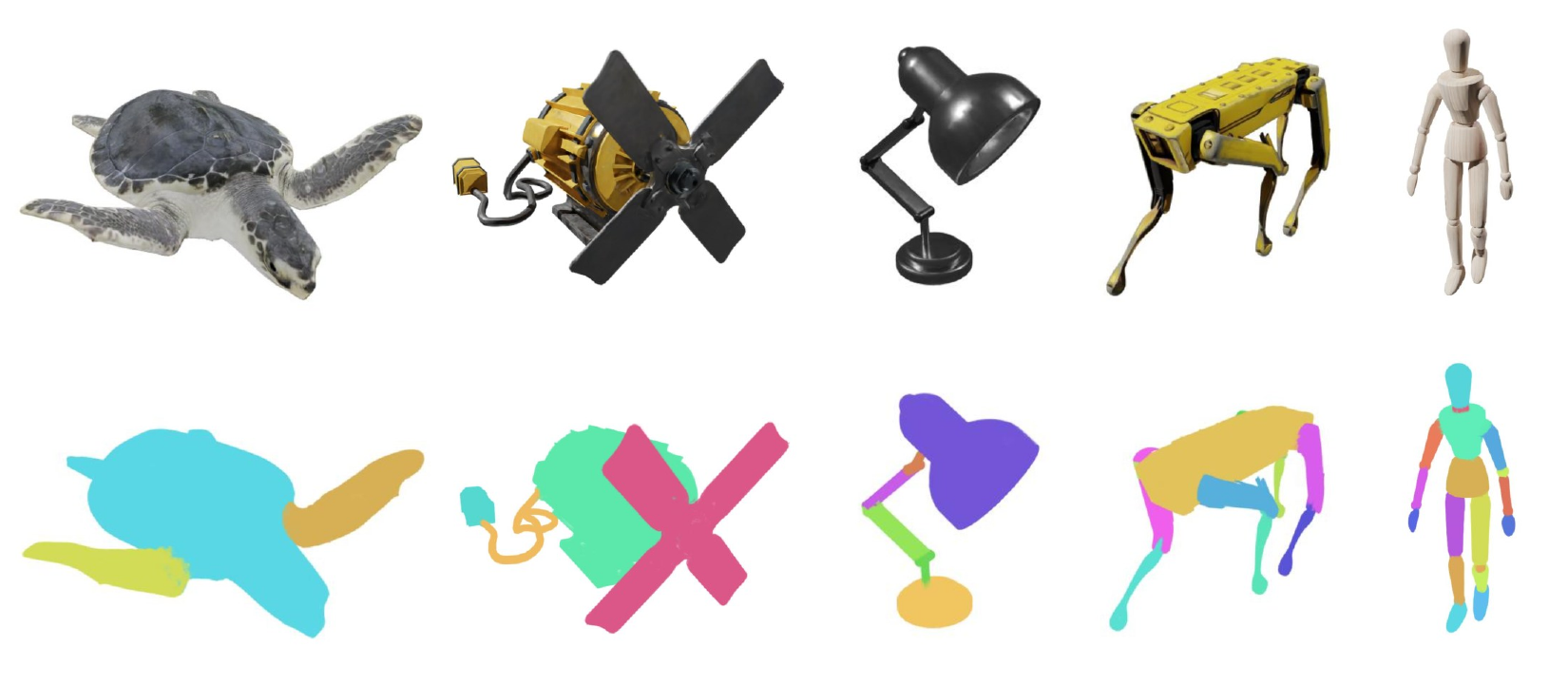}
    \caption{
    \textbf{Synthetic objects} used for quantitative evaluation.
    (Top) RGB renderings, (Bottom) part segmentations.
    }
    \label{fig:synth}
    \vspace{-1em}
\end{figure}

\subsection{Predict}
\label{sec:predict}
\textbf{Feed-Forward Model}
\algabbr{} learns a per-frame pose prediction model, which takes the RGB frame as input 
and outputs the part configuration and the camera-to-object transform; this model is continually updated over each cycle iteration. \algabbr{} uses a lightweight transformer-decoder on top of frozen DINOv2 features to predict the object configuration from an image. Importantly, the model explicitly disentangles object pose and part poses in its outputs, which allows the part configurations to be used for quasi-multiview supervision (Sec.~\ref{sec:optimize}).

To initialize the predictive model, we run RSRD~\cite{rsrd} on the input video, which provides a rough estimate of the part pose trajectory. Often this contains failures due to occlusions, so we train the predictive model on output poses from RSRD along with the identity object poses.

\textbf{Training} See Sec.~\ref{sec:distill} for a detailed explanation of synthetic data generation. During training, we augment input images with color jitter and random masking to reduce the distribution shift from synthetic renders to real videos with partial occlusions or under different illumination conditions. After each cycle iteration, we continue fine-tuning the predictive model on newly generated synthetic data.



Notably, \algabbr{} is robust to failure cases in video tracking because it is trained on synthetic image–pose pairs, where some poses may be incorrect. Synthetic data provides a diverse set of image-pose pairs, so \algabbr{} effectively learns a reliable mapping from images to poses. Good synthetic images align with real images, reinforcing correct predictions, while bad synthetic images remain out of distribution, ensuring they do not degrade inference quality. 



\subsection{Optimize}
\label{sec:optimize}
\algabbr{} 
refines the per-frame pose estimates
through global trajectory optimization.
This optimization uses inverse rendering to backpropagate pixel-based losses to pose estimates, and leverages 3D/temporal regularization along with a multiview mining technique to improve global coherence. Here an advantage of long-form input video becomes clear: as the video observes object motion multiple times from different angles, \algabbr{} estimates frame correspondences for quasi-multiview supervision using the part configurations predicted for each frame from the feed-forward model.
Poses for each frame are initialized by passing video frames into the latest feed-forward model after masking the object. Optimization proceeds in minibatches of 20 frames each over 50 epochs to optimize both part configuration poses and object pose.

\noindent \textbf{Optimization Losses} The part poses are then refined through an optimization process guided by multiple loss functions. 
Specifically, given an image \( I_i \), the predicted part poses \( \text{T}_{parts}^{obj} \), and the predicted camera pose \( \text{T}_{obj}^{cam} \), we optimize the following losses.




\textit{DINO Feature Loss:} To ensure consistency between the predicted part poses and the input image features, we minimize the feature distance in the DINO feature space: $\mathcal{L}_{\text{DINO}} = \left\| F_{\text{DINO}}(I_i) - R_{\text{DINO}}(\text{T}_{obj}^{cam} \times \text{T}_{parts}^{obj}) \right\|^2$, where $F_{\text{DINO}}(\cdot)$ represents the frozen DINOv2 encoder, and $R_{\text{DINO}}(\cdot)$ denotes the rendered DINO feature. 

\textit{Relative Depth Loss:}  
To encourage depth consistency, we minimize the discrepancy between the depth predicted by a DepthAnything~\cite{depthanything} and the rendered depth image. As depth predictions are non-metric, we use the pair-wise ranking loss proposed in SparseNeRF~\cite{wang2023sparsenerf}, sampled over 10,000 pairs across the image object mask.



\textit{Mask Loss:}  
To align the predicted object shape with the input segmentation mask, we compute the pixel-wise MSE between rendered opacity and image mask:
$
    \mathcal{L}_{\text{mask}} = \left\| F_{\text{mask}}(I_i) \  - R_{\text{mask}}(\text{T}_{obj}^{cam} \times \text{T}_{parts}^{obj})\right\|^2,
$
where \( F_{\text{mask}}(\cdot) \) represents the ground-truth object mask from Segment Anything v2~\cite{ravi2024sam2}, and \( R_{\text{mask}}(\text{T}_{obj}^{cam} \times \text{T}_{parts}^{obj}) \) denotes the rendered accumulated opacity.

\textit{Static Prior:} 
To aid optimization in converging to 3D-coherent part poses, we impose a static prior which penalizes parts for drifting away from their initial relative configuration. First we find neighboring parts based on how many gaussians are within a threshold distance from each other. The centroid of the connection area between parts \( a \) and \( b \) is denoted as \( c \). To regularize the relative transformations between parts \( a \) and \( b \), we track the motion of \( c \) in each part's local coordinate frame and add a loss encouraging them to consistently estimate the same point relative to the object frame, denoted \( o \).
Specifically, over all connected pairs we minimize:
$
    \mathcal{L}_{\text{static}} = \sum_{a,b} 
        \left\| T_{a}^o T_{c}^a - T_{b}^o T_{c}^b \right\|^2
$





\textit{Temporal Smoothing:} During optimization, we encourage temporal smoothness by computing each frame's velocity with a 3-point finite difference and penalizing the difference between the current frame's velocity and the average of its neighbors. This loss is applied only on part poses sampled within a given minibatch.

\begin{figure}[t!]
    \includegraphics[width=\linewidth,trim=4 4 4 4,clip]{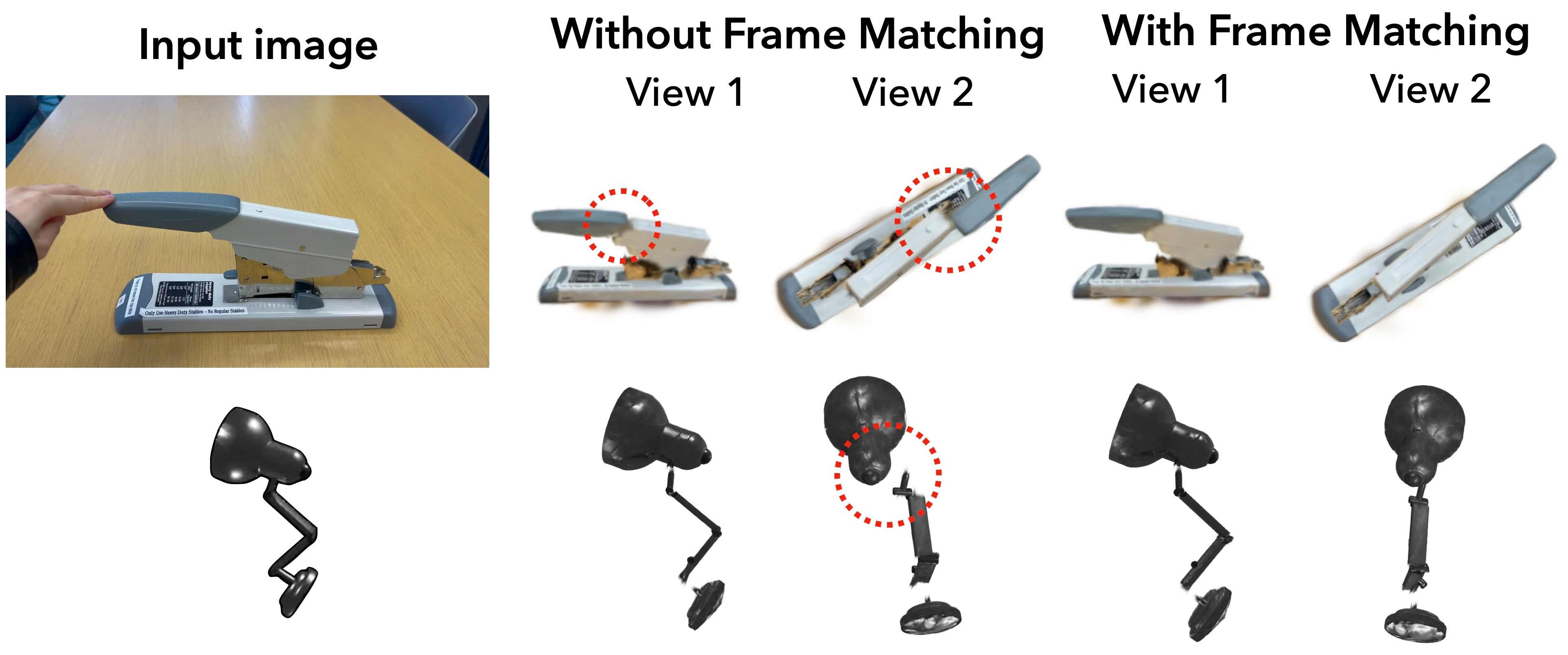}
    \captionof{figure}{
    \textbf{Quasi-multiview ablation.}
    While \algname{} without quasi-multiview generally avoids catastrophic failures, temporal consistency and photometric losses alone fail to correct inter-part pose errors due to depth ambiguity, causing poses to appear disconnected from novel viewpoints (red circles). The quasi-multiview signal from frame matching effectively resolves these errors.}
    \label{fig:matching}
    \vspace{-1em}
\end{figure}

\textbf{Quasi-Multiview Supervision}
To effectively take advantage of a video observation of repeated motions, we use the output of the pose prediction model to find another frame correspondences where the local part configuration is similar, and use it as a quasi-multiview frame to supervise the optimization process. 
Given the frame that is closest in local pose configuration, 
we then render views from \textit{both} perspectives to optimize the same object configuration, which distills multi-view information into the parameters by forcing it to explain both frames simultaneously. For each video frame, we predict the top $\frac{N}{4}$ neighbors as the candidate matching frames based on average SE(3) distance of parts in object frame, filtering neighbors with non-max suppression to leave only the modes. 
During mini-batch optimization, for each frame, we sample an additional matching frame from this set to compute pixel-based losses from another perspective. For the set of candidate matching frames, we first compute the part configuration similarity and camera distance relative to reference frame. 
Then, the matching frame is sampled using importance sampling based on the normalized camera distance, facilitating the acquisition of more diverse viewpoint information. We weight the loss of the matching frame proportionally to the normalized part configuration similarity. This multiview supervision helps resolve depth ambiguity errors, as shown in Fig.~\ref{fig:matching}.




\subsection{Distill}
\label{sec:distill}
After improving the predicted part pose initialization via optimization, the final step of the cycle is to fit the predictive model to the new poses. To do this we generate a large dataset of novel views of the tracked trajectory from camera viewpoints with complete 360$^\circ$ coverage and label them with the ground-truth part and object poses used to render these images. Re-distilling on synthetic has two key benefits: 1) We can mine viewpoints unseen in the original video which makes it less prone to overfitting on the input video frames. 2) Re-distillation improves the accuracy of the predicted results by updating with the latest data from optimization outputs.

\textbf{Synthetic Data Generation} 
To train the pose prediction model, we generate a sufficient number of images paired with corresponding part and camera poses. Two camera sampling methods are employed for generating the training camera views: one involves perturbing poses along a hemispherical facing the object, while the other samples new camera views near the optimized camera view from the previous loop. For the first loop, we use only the hemisphere method to generate camera views, as the RSRD initialized camera views tend to be noisy. In subsequent iterations, both camera sampling strategies are utilized, enabling the model to refine its predictions while maintaining diversity by incorporating new random views. To sample random camera views, we generate camera poses along a hemisphere with a radius determined by the initial camera-to-object distance $d_{co}$. The radius range is $[0.5 \cdot d_{co}, d_{co}]$, centered around the object (with the dataset scaled so that cameras fit within a unit cube). We grid-sample 600 azimuthal and 30 elevation angles to get a total of 18,000 viewpoints, and randomly perturb their local tilt and pitch $\pm20^\circ$, selecting a random frame of the trajectory to render. This results in a highly diverse set of viewpoints of the object trajectory. To sample new camera views near the optimized camera view, we generate augmented camera poses by adding random transformations on the optimized camera poses. The augmented cameras and the optimized camera poses are randomly paired with the optimized object part poses to generate an augmented dataset. We also upsample the optimized camera poses and object part poses pairs to the same amount of the augmented dataset.

\begin{figure*}[t!]
    \includegraphics[width=.98\linewidth,trim=4 4 4 4,clip]{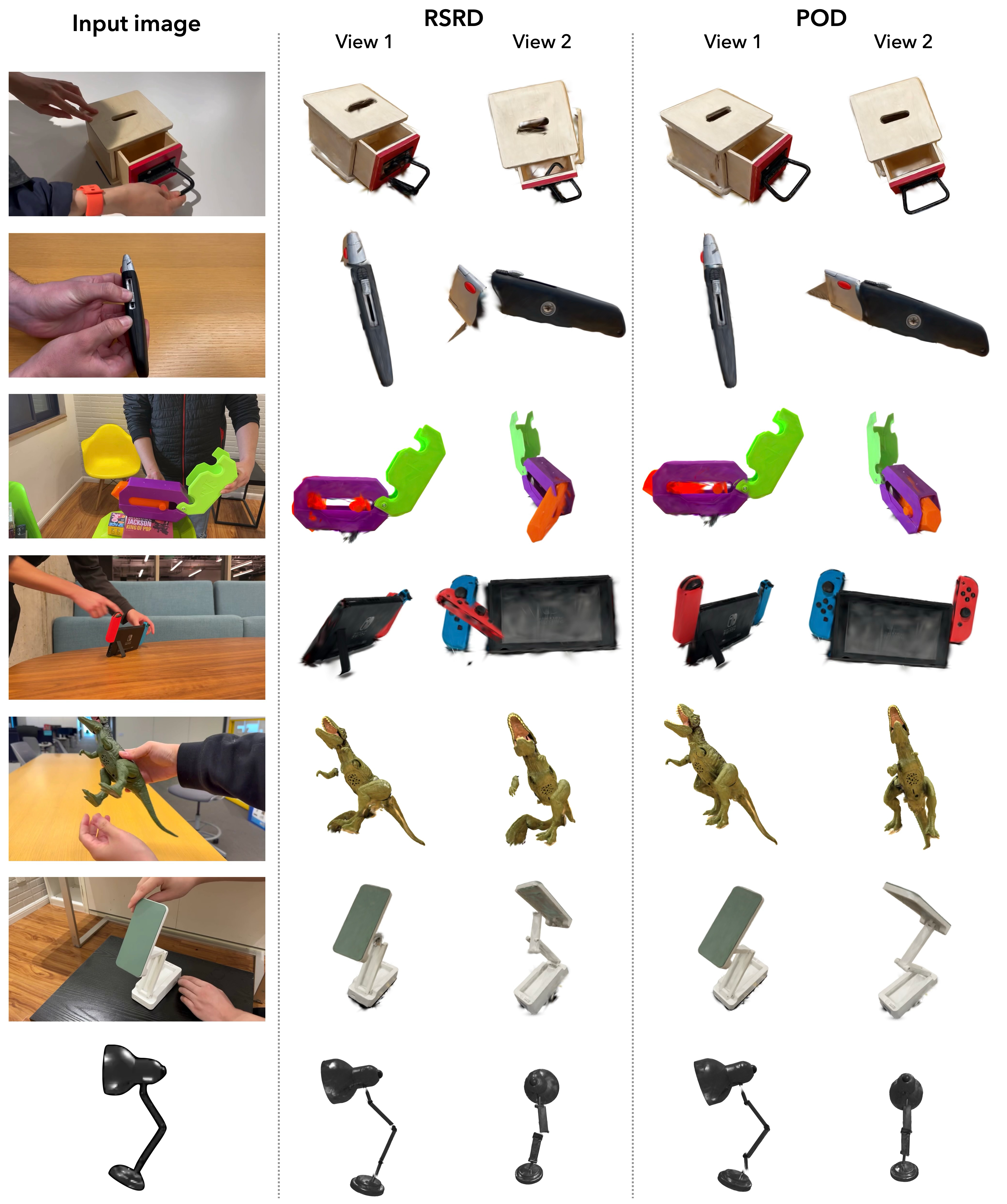}
    \caption{
    \textbf{Results.}
    We compare \algname{}'s feed-forward model with outputs from RSRD~\cite{rsrd}, an optimization-only baseline. RSRD often produces incoherent solutions under significant occlusion (e.g., intersecting Nintendo Switch components) or produces poses that seem plausible from the input viewpoint but break from novel viewpoints due to slight depth and pixel mismatches. In contrast, \algname{} predicts more 3D-consistent configurations by leveraging frame matching and synthetic data distillation (see Sec~\ref{sec:results_qualitative} for details).
    }
    \label{fig:results}
\end{figure*}




\begin{figure}[t]
    \centering
    \includegraphics[width=\linewidth]{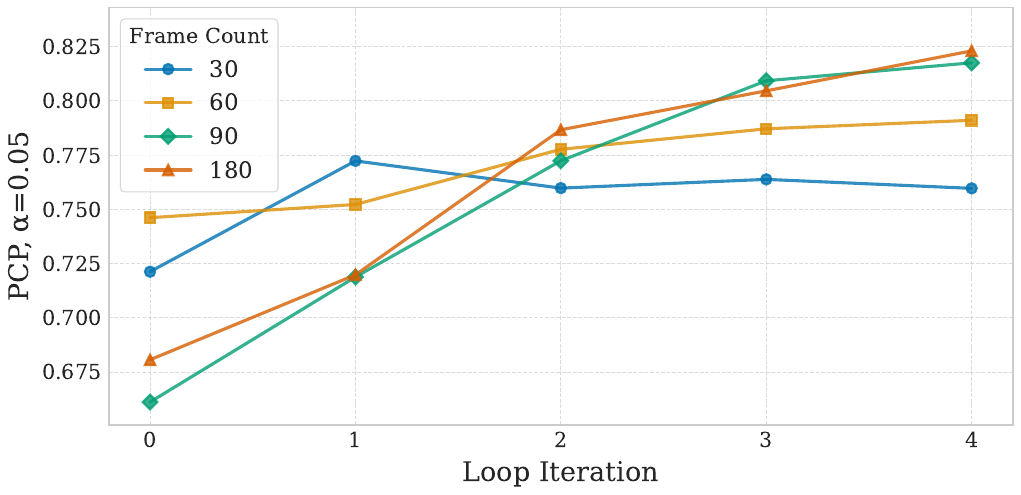}
    \caption{\textbf{Prediction accuracy versus loop iteration}: We report synthetic ground truth evaluation averaged over 5 objects across each loop iteration with different length input sequences of the same motion at 30 fps (Sec.~\ref{sec:results_synthetic}). Each loop improves the accuracy of predictions owing to its self-distillation and optimization. 
    \algabbr{}'s performance improves in tracking accuracy on longer sequences which benefit more from quasi multi-view supervision. Longer videos also benefit more from successive cycle iterations.}
    \label{fig:longer_video}
    \vspace{-1em}
\end{figure}


\begin{table}[t]
  \centering
  \resizebox{0.5\textwidth}{!}{  
  \resizebox{0.5\textwidth}{!}{  
  \begin{tabular}{l c c c c}
    \toprule
    Method & MSE & PCP $\alpha=0.05$ & PCP $\alpha=0.04$ & PCP $\alpha=0.03$ \\
    \midrule
    \algabbr{} - View Aug &0.0465 &0.752 &0.674 &0.561 \\
    \algabbr{} - RSRD Init &0.0434 &0.760 &0.696 &0.603\\
    \algabbr{} - Multiview &0.0464 &0.759 &0.683  &0.570\\
    \midrule
    RSRD~\cite{rsrd} &0.0952 &0.454 &0.368 & 0.266\\
    POD & \textbf{0.0422} & \textbf{0.778} & \textbf{0.714} & \textbf{0.622}\\
    \bottomrule
  \end{tabular}
  }
  }
  \caption{\textbf{Quantitative Evaluation} (Sec~\ref{sec:results_synthetic}). The top section shows ablations of our method, and the bottom compares \algabbr{} with a pure optimization baseline.}
  \label{tab:ablations_comparison}
  \vspace{-1em}
\end{table}



\section{Experiments}
We evaluate \algabbr{} on both synthetic and real-world data, totalling 19 objects, 14 real and 5 synthetic. These objects range from common household (scissors) to highly long-tail like art sculptures or toy figurines. This highlights \algabbr{}'s ability to ingest an object scan and bootstrap its own understanding of its motion. Our demonstration videos range from 10 seconds to up to 30 seconds long, and are captured with a mostly static camera and a person manipulating the object at different orientations. Please see the supplemental for implementation details. 

\subsection{Qualitative Results}
\label{sec:results_qualitative}
\algabbr{}'s results are best viewed in the accompanying videos. Fig~\ref{fig:results} shows selected frames from the input video sequence, comparing RSRD outputs to the output of \algabbr{}'s feed-forward model after 5 cycle iterations. Despite only seeing one frame, \algabbr{}'s predictive model can predict reasonable 3D-consistent part configurations which surpass the global optimization of RSRD. Notably, \algabbr{} can predict coherent 3D part configurations despite heavy self-occlusion of the input (Switch, knife, lamp), and more reliably resolves depth ambiguity with quasi multi-view (drawer handle, colorful blade). This tolerance of occlusion is especially apparent in videos, where frequently RSRD failes irrecoverably. This ability is largely in part to supervising with synthetic data augmented with random masking. In addition, though synthetic evaluation shows a moderate difference in PCP metrics, these small errors are quite visually jarring when viewed from certain viewpoints (bottom row lamp, the hinge disconnects).

\subsection{Synthetic Ground Truth Evaluation}
\label{sec:results_synthetic}
To numerically evaluate the performance of our method, we conduct experiments on synthetic ground truth data by selecting 5 animatable objects (Fig.~\ref{fig:synth}) from Sketchfab and rendering a 6 second spiral trajectory around the object while the animation loops, similar to how Objaverse~\cite{deitke2023objaverse} extracts moving assets. Each video is 6 seconds long, and the camera trajectory consists of a spiral which loops every 2 seconds. Importantly, this synthetic data is \textit{not} synchronized multiview data like many synthetic datasets, and though the data is easier in that there are no occlusions or pauses, it mimics monocular nature of our real-world captured data. To evaluate the performance of \algabbr{}, we subsample gaussian points and project them onto the mesh's barycentric coordinates. During evaluation, we evaluate mean squared error (MSE) and percentage of correct points (PCP) on the ground truth barycentric interpolated points versus their counterparts in the reconstructed 3DGS model.

\textbf{Overall Results} Tab.~\ref{tab:ablations_comparison} summarizes our results. 
We compare against an optimization-only baseline in RSRD~\cite{rsrd}, improved by adding our mask loss which we observe and prior work has shown aids in tracking robustness with 3DGS models~\cite{yu2025pogs,zhang2024egogaussian}. 
\algabbr{} drastically outperforms RSRD~\cite{rsrd}, an optimization-only baseline, which frequently irrecoverably fails on long input videos. In contrast, \algabbr{} is able to recover and improve on those failures because of its iterative synthetic data distillation cycle. We present 3 different thresholds for PCP, and note significantly more robustness to tightened accuracy tolerance in \algabbr{}.

\textbf{``Do longer videos aid reconstruction?"} We conduct an experiment on synthetic data where we vary the length of input video from 1 second to 6 seconds, truncating the observed motion at different points. We evaluate PCP on the same 30 frames across all video lengths to keep comparisons fair. We observe that longer videos yield better prediction accuracy, causing a performance delta of 6\%. Also note that at loop 0 longer video's performance suffers compared to short videos as optimization struggles to reconstruct motion accurately (Fig.~\ref{fig:longer_video}). 

\textbf{``To what extent do successive cycles improve?"} To answer this question we plot ground truth evaluation results over each loop iteration, and observe a clear improvement as the cycle progresses (Fig.~\ref{fig:longer_video}), in the longest video improving PCP by 14\%. This effect is less pronounced for short videos where optimization poses little benefit over predicted poses. As \algabbr{} sees more frames, it can better leverage the quasi-multiview supervision from longer videos, leading to better performance over iterations. 

\begin{figure}[t!]
    \centering
    \includegraphics[width=\linewidth]{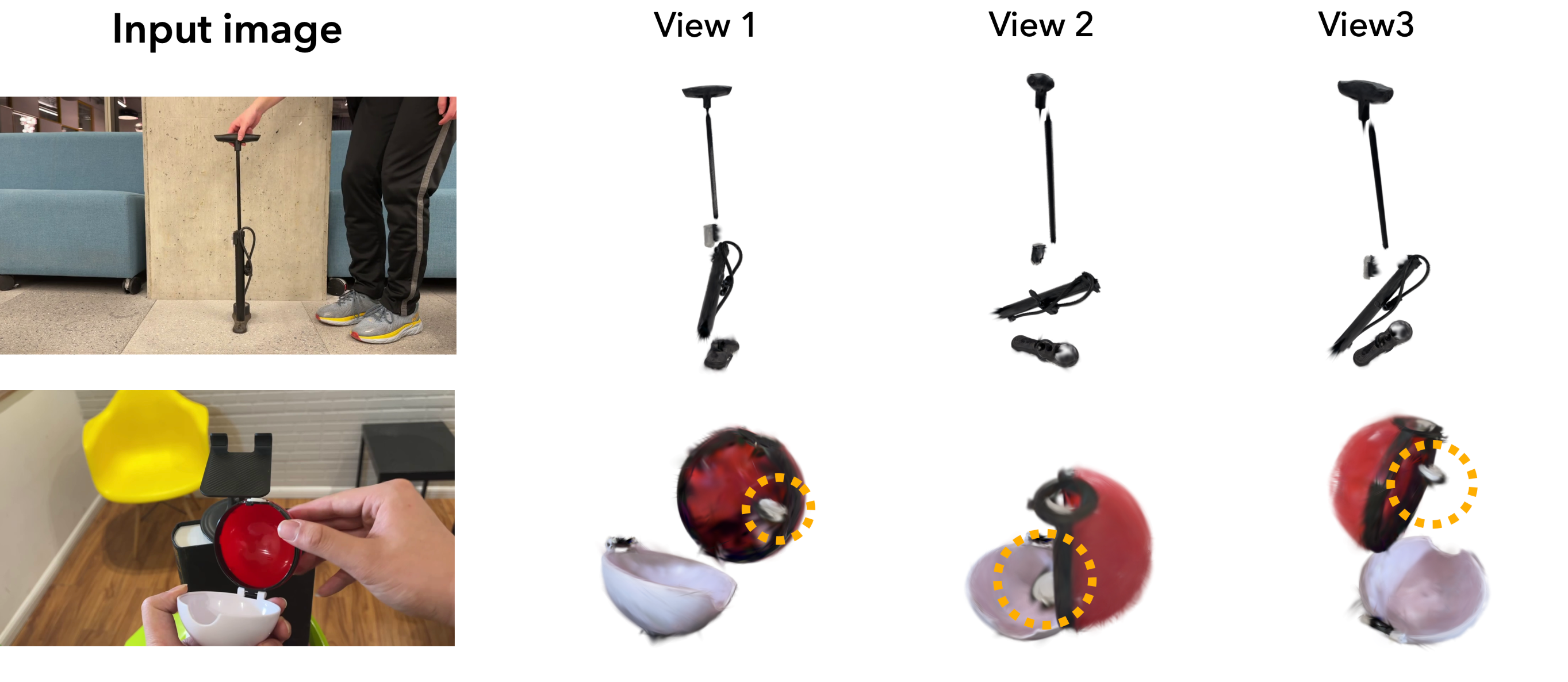}
    \caption{
    \textbf{Failure cases.}
    \algname{} may struggle to track thin (top) or small (bottom) parts, which are especially sensitive to noise in DINOv2 features. Rotationally symmetric parts can also be a challenge due to pose ambiguity.
    }
    \label{fig:pod-failures}
    \vspace{-1em}
\end{figure}

\subsection{Ablations}
We separately ablate the synthetic view generation, RSRD initialization, and quasi-multivew supervision of \algabbr{}. Table~\ref{tab:ablations_comparison} shows ablations of our method. Without view augmentation, we distill the predictive model only on viewpoints matching the input video, which causes an overall accuracy dropoff of  as the predictive model has less opportunity to resolve depth ambiguity in its training data. Without RSRD initialization, \algabbr{} initializes its pose predictor by training on novel views of the static object configurations. We note the fact that without this initialization \algabbr{} is still able to converge over loop iterations to almost the same performance within 5 loops. Without quasi-multiview, \algabbr{} performance suffers as its optimization phase struggles to refine depth ambiguity.

\section{Limitations and Future Work}
One limitation of \algabbr{} is the need to retrain a predictor for each new object. An exciting direction for future work is to train a single predictive model across multiple objects and videos. This requires designing an architecture capable of handling varying numbers and functions of object parts. In addition, \algabbr{} is partially reliant on the quality of 3D object segmentation -- if a part is not segmented it is impossible to track it over time. \algabbr{} implicitly assumes that motions are repeated at least once throughout the input video, or quasi-multiview supervision would fail to produce correct correspondences. It also struggles with objects that have thin or very small parts, as they can be torn apart by small amounts of noise in the computed DINOv2 features, or objects with rotational symmetry which yield ambiguous outputs for the predictive model, as shown in Figure~\ref{fig:pod-failures}. Finally, future work could explore a generative model to modernize the architecture with conditional diffusion, 
 and convert the model to a simulator-compatible format.

\section*{Acknowledgment}
{\small
\noindent\raggedright
This project was performed at UC Berkeley in affiliation with the Berkeley AI Research (BAIR) Lab. This project is supported in part by {NSF:CNS-2235013}, Bakar Fellows, and BAIR Sponsors.
}
{
    \small
    \bibliographystyle{ieeenat_fullname}
    \bibliography{main}
}

\clearpage
\appendix
\setcounter{page}{1}
\maketitlesupplementary

\section{List of objects}


We include the list of real objects that were used for qualitative results; all of them were captured using smart phone camera as well as the long monocular videos. A sample of objects and their corresponding groups are shown in the main paper. 

The full list of objects and their names are:
\begin{enumerate}[\hspace{0.5cm}1.]
    \item Tractor
    \item Redbox
    \item Stapler
    \item Scissors
    \item Retractable knife
    \item Folded lamp
    \item Carrot knife
    \item Vacuum
    \item Barbarian
    \item Switch
    \item T-Rex
    \item Pokeball
    \item Wooden Drawer
    \item Bike Pump
\end{enumerate}

\section{POD Implementation Details}
\subsection{Predictive Model}
\algabbr{}'s predictor extracts DINOv2 features from the input image, positionally encodes them with a sinusoidal embedding, and uses a 3-layer transformer decoder to decode 2 output tokens. The first token is passed through an MLP to predict object to camera pose, and the second is passed into a distinct MLP to predict a vector of all part poses. We represent neural network SE(3) outputs with the 6DoF Gram-Schmidt representation for SO(3) and a 3-vector for position. For the first cycle, we train the predictor model for 250 epochs on the synthetic data with a batch size of 1600 using L1 loss on its outputs. The predictor model is finetuned for 150 epochs in the following cycles. During inference, to match the training distribution of synthetic data we mask the object with SAMv2~\cite{sam2}. 

\subsection{Optimization}
Before batch optimization, we optimize object's global transformation for 15 steps for each frame while keeping the part poses fixed, to better align the rendered image with the ground truth signal. With the optimized global transformation, we combine it with the predicted camera pose to obtain the optimized camera, which serves as the initial camera pose for batch optimization.
Temporal smoothness losses alone do not completely remove temporal discontinuities in high dimensional pose space without data-driven priors~\cite{huang2017towards,ye2023slahmr}, hence we further apply DCT smoothing on our results (both ours and baselines) to remove high frequency jitter artifacts in the final results, following the DCT low-pass filtering method proposed in~\citet{akhter2008nonrigid}.

\begin{figure}
    \centering
    \includegraphics[width=\linewidth]{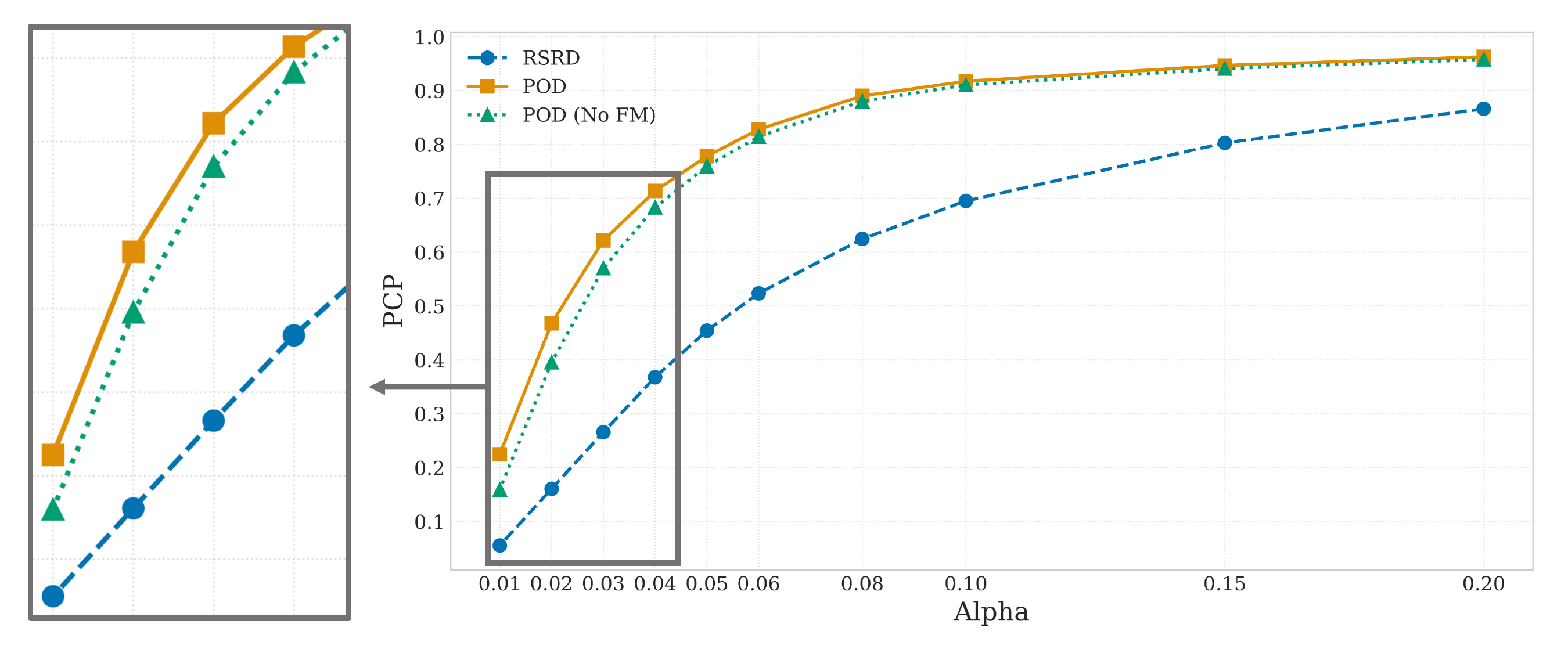}
    \caption{Exhaustive results for PCP thresholds. POD improves over RSRD in accuracy over all thresholds. We also compare against no frame matching (quasi-multiview), which results in worse performance at more strict thresholds where minor deviations are more noticable.}
    \label{fig:enter-label}
\end{figure}

\subsection{PCP Metric}
We compute PCP by measuring the percentage of points within a threshold of the correct corresponding point. A tighter threshold will yield more strict tolerance on error. We show exhaustive results for PCP at different thresholds here.

\end{document}